\def\year{2020}\relax
\documentclass[letterpaper]{article} % DO NOT CHANGE THIS
\usepackage{aaai20}  % DO NOT CHANGE THIS
\usepackage{times}  % DO NOT CHANGE THIS
\usepackage{helvet} % DO NOT CHANGE THIS
\usepackage{courier}  % DO NOT CHANGE THIS
\usepackage[hyphens]{url}  % DO NOT CHANGE THIS
\usepackage{graphicx} % DO NOT CHANGE THIS
\usepackage{graphicx}  % DO NOT CHANGE THIS

\usepackage{amsmath}
\usepackage{diagbox}
\usepackage{multirow}
\usepackage{subcaption}
\usepackage{booktabs}
\usepackage{xcolor}
\usepackage{tabularx}
\usepackage{etoolbox,siunitx}
\usepackage{xspace}
\usepackage{xifthen}
\usepackage[utf8]{inputenc}

\usepackage{relsize}
\usepackage[tracking=smallcaps]{microtype}
\newcommand\relsc[1]{{\relscale{1.09}\textsc{#1}}}
\SetTracking{encoding=*, shape=sc}{25}

\robustify{\bfseries}
\newcommand{\citet}[1]{\citeauthor{#1} \shortcite{#1}}\newcommand{\citep}{\cite}\newcommand{\citealp}[1]{\citeauthor{#1} \citeyear{#1}}

\newcommand\DSTC[1][]{\mbox{\relsc{dstc}\ifthenelse{\isempty{#1}}{}{-#1}}\xspace}
\newcommand\dstc[1][]{\mbox{\relsc{dstc}\ifthenelse{\isempty{#1}}{}{-#1}}\xspace}
\newcommand\metalwoz{\mbox{\relsc meta\relsc{l\!w\hspace{-1pt}o}z}\xspace}
\newcommand\multiwoz{\relsc multi\relsc{w\hspace{-1pt}oz}\xspace}
\newcommand{\GPT}[1][]{\relsc{gpt}\ifthenelse{\isempty{#1}}{}{-#1}\xspace}
\newcommand\LSTM{\relsc{lstm}\xspace}
\newcommand\GRU{\relsc{gru}\xspace}
\newcommand\BERT{\relsc{bert}\xspace}
\newcommand\NLU{\relsc{nlu}\xspace}
\newcommand\METEOR{\relsc{meteor}\xspace}
\newcommand\BLEU[1][]{\relsc{bleu}\ifthenelse{\isempty{#1}}{}{-#1}\xspace}
\newcommand\CIDEr{\relsc{cide}r\xspace}
\newcommand\ROUGEL{\relsc{rouge-l}\xspace}
\newcommand\HRED{\relsc{hred}\xspace}
\newcommand\NLGEval{\relsc{nlge}val\xspace}
\newcommand\ZSDG{\relsc{zsdg}\xspace}
\newcommand\ELMo{\relsc{elm}o\xspace}
\newcommand\WOz{\relsc{wo}z\xspace}
\newcommand\SPFT{\relsc{sp}+\relsc{ft}\xspace}

\title{Hybrid Generative-Retrieval Transformers for Dialogue Domain Adaptation}
\author{
  Igor Shalyminov\thanks{Work done during an internship at Microsoft Research Montréal.}\\ % All authors must be in the same font size and format. Use \Large and \textbf to achieve this result when breaking a line
  Heriot-Watt University\\ %If you have multiple authors and multiple affiliations
  Edinburgh, UK\\
  is33@hw.ac.uk % email address must be in roman text type, not monospace or sans serif
  \And 
  Alessandro Sordoni, Adam Atkinson, Hannes Schulz\\
  Microsoft Research\\
  Montréal, Canada\\
  \{alsordon,adatkins,haschulz\}@microsoft.com
}
\begin{document}

\maketitle

\begin{abstract}
% 1000 characters max - currently at 990 according to Google Docs
Domain adaptation has recently become a key problem in dialogue systems research. Deep learning, while being the preferred technique for modeling such systems, works best given massive training data. However, in the real-world scenario, such resources aren't available for every new domain, so the ability to train with a few dialogue examples can be considered essential.
Pre-training on large data sources and adapting to the target data has become the standard method for few-shot problems within the deep learning framework.
In this paper, we present the winning entry at the fast domain adaptation task of \dstc[8], a hybrid generative-retrieval model based on \GPT[2] fine-tuned to the multi-domain \metalwoz dataset\footnote{Code is publicly available at \url{http://tiny.cc/grtr}.}.
Robust and diverse in response generation, our model uses retrieval logic as a fallback, being SoTA on \metalwoz in human evaluation ($>$4\% improvement over the 2nd place system) and attaining competitive generalization performance in adaptation to the unseen \multiwoz dataset.
\end{abstract}

\section{Introduction}
\label{sec:intro}
Goal-oriented dialogue is an area of increasingly high interest, both from academic and industrial perspectives. Data-driven approaches to developing such systems \cite{Lemon:2012:DMA:2412075} proved to be more flexible and scalable to various scenarios and domains than previous techniques widely employed in industry, mostly based on expert knowledge. The benefits of methods based on machine learning (especially deep learning) can only be experienced when there are excess amounts of training data available; however, in real-world scenarios, there's only a small amount of initial data available for a new domain. Training techniques must make the most of this small data, i.e. work in a \textit{data-efficient} way, in order to enable rapid development of dialogue models for an ever-increasing number of domains and tasks. The most promising method to achieve this under the deep learning framework has become transfer learning where a large, generic model is first trained from a highly represented source of data, after which it gets adapted to the target task. 
%(see Section~\ref{sec:related} for examples).

In this paper, we explore this problem through the Eighth Dialogue System Technology Challenge (\dstc), Fast Domain Adaptation task. Specifically, we propose a hybrid generative/retrieval dialogue model leveraging knowledge transfer from a large-scale pre-trained general-purpose language model. Our model is able to maintain goal-oriented dialogue in a closed domain having only been exposed to a small set of in-domain dialogues as the domain description. Our hybrid model achieves state-of-the-art performance on the \metalwoz dataset when evaluated with human judges, and attains competitive generalization level in adapting to goal-oriented \multiwoz dataset unseen at the main training stage. Automated word overlap-based metrics demonstrate that it outperforms a series of competitive baselines---both generative-only and retrieval-only models.

\section{Related work}
\label{sec:related}

\begin{figure*}[ht!]
  \centering
  \includegraphics[width=0.95\linewidth]{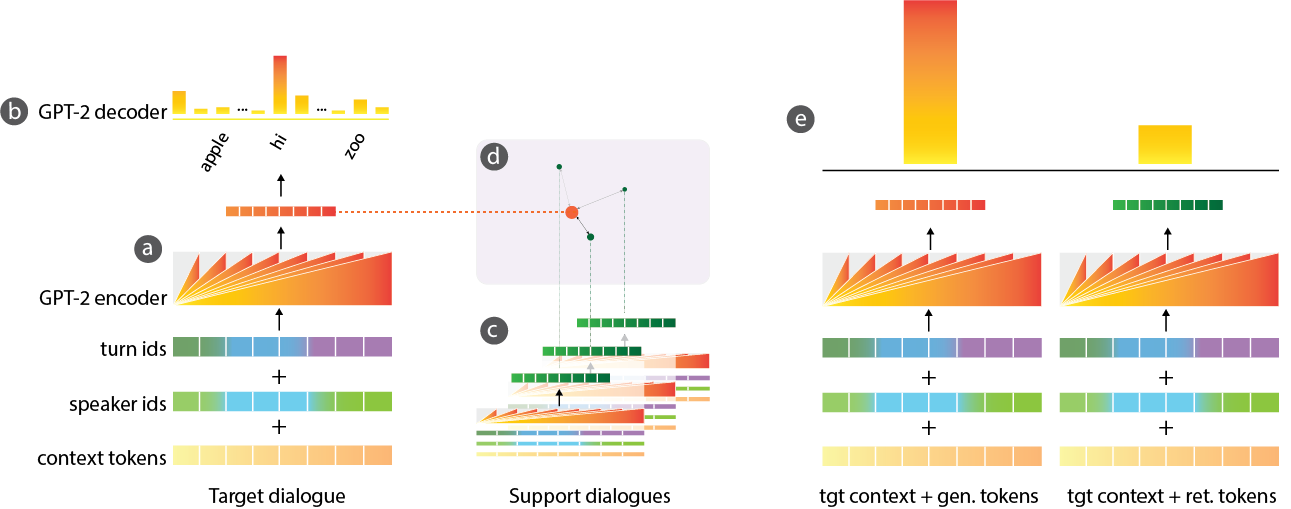}
  \caption{Model diagram: (a) encode target the dialogue context and (b) produce the `generated candidate'; (c) encode support dialogue contexts in a similar way; (d) find the nearest `support' neighbor and select its response as the `retrieved candidate'; (e) finally, rank the two candidates given the target context and produce the final result.}
  \label{fig:model}
\end{figure*}

Generative dialogue modeling is an actively researched area, with the sequence-to-sequence (seq2seq) model \cite{DBLP:journals/corr/VinyalsL15} gaining wide adoption in both chat-oriented \cite{DBLP:conf/aaai/SerbanSBCP16} and goal-oriented dialogue \cite{DBLP:conf/sigdial/ZhaoLLE17}. Initially these architectures were based on Recurrent Neural Networks such as \LSTM \cite{Hochreiter:1997:LSM:1246443.1246450} or \GRU \cite{DBLP:journals/corr/ChungGCB14} which were quite challenging to train on large amounts on conversational data, causing researchers to focus on improving response diversity \cite{Li2015ADO} and the overall dialogue consistency \cite{Li2016NeuralNM}. Quite recently, self-attention mechanisms, like those used in the Transformer \cite{DBLP:conf/nips/VaswaniSPUJGKP17}, have been adopted for conversation models---together with large-scale pre-training, it resulted in a new generation of seq2seq architectures. % (they will be discussed later).

The data efficiency of dialogue systems has also been extensively researched in the past. Initially, when modular dialogue system architecture was the prevalent approach, dialogue managers and state trackers were the components that data-efficient methods were applied to the most. As such, the dialogue state tracker domain adaptation task was initially proposed in \DSTC[3] \cite{DBLP:conf/slt/HendersonTW14}~--- that challenge featured approaches like Bayesian Processes \cite{DBLP:journals/csl/GasicMRSUVWY17} and Recurrent Neural Networks \cite{DBLP:conf/acl/MrksicSTGSVWY15}. Later research was focused on data-efficiency of dialogue managers, for instance \citet{DBLP:conf/acl/WilliamsAZ17} introduced a model designed for bootstrapping from limited training data and further fine-tuning in the reinforcement learning fashion. Furthermore, a recent paper by \citet{DBLP:journals/corr/abs-1811-11707} proposed a dialogue management model which used a unified embedding space for user and system turns allowing for efficient cross-domain knowledge transfer.

End-to-end dialogue response generation, the technique that followed modular architectures with the arrival of large conversational datasets, was also eventually approached in a data-efficient way. One such method used prior linguistic knowledge to improve zero-shot performance: \citet{Eshghi.etal17a} proposed a linguistically informed model based on an incremental semantic parser \cite{Eshghi.etal11} combined with a reinforcement learning-based agent. The parser was used for both maintaining the agent's state and pruning the agent's incremental, word-level generation actions to those leading to syntactically correct word sequences. While outperforming end-to-end dialogue models on bAbI Dialog Tasks \cite{babi} in the extreme zero-shot case \cite{Shalyminov.etal17}, this method inherited the limitations of the dialogue grammar --- specifically, it is limited to a single closed domain until a wide-coverage grammar is available.

\citet{DBLP:conf/sigdial/ZhaoE18} introduced zero-shot dialogue generation (\ZSDG) framework under which a dialogue system was trained on dialogues from several source domains and a small amount of annotated utterances from the target domain. The key feature in their framework was the unified latent space which was used to encode user's queries, dialogue contexts, and annotations.

Later, Shalyminov et\,\,al. (\citeyear{DBLP:journals/corr/abs-1908-05854,DBLP:journals/corr/abs-1910-01302}) proposed Dialogue Knowledge Transfer Networks which approached the problem in a few-shot setup with a separate out-of-domain pre-training stage on a large goal-oriented corpus (\metalwoz, \citealp{lee2019multi-domain}). In those approaches, \metalwoz was used as source dataset for transfer, whereas we treat it as the target dataset. While the authors used full target-domain dialogues, they ended up using only a fraction of \ZSDG's data in terms of the number of utterances.

More generally, transfer learning has been widely adopted for natural language problems with the emergence of large-scale pre-trained text representation models like \ELMo \citep{DBLP:conf/naacl/PetersNIGCLZ18}, \BERT \citep{DBLP:conf/naacl/DevlinCLT19}, and \GPT[2] \citep{gpt2}. When applied to dialogue response generation, the most successful approaches made use of a Transformer for chat-oriented dialogue \citep{DBLP:journals/corr/abs-1901-08149} and \GPT/\mbox{\GPT[2]} for goal-oriented dialogue \citep{DBLP:journals/corr/abs-1907-05774}. Our approach is based on a similar technique, though in addition to fine-tuning a pre-trained model to our task, we augment the generative model with a retrieval component in a hybrid approach.

Finally, another recent approach applied to the problem of few-shot dialogue generation is meta-learning \citep{DBLP:conf/acl/QianY19}, under which the task is split into multiple subtasks corresponding to dialogue domains. For each of them, a specialized dialogue model was trained, with their training progress then merged into the main model. In general, the intuition behind meta-learning is training a base model which would be best suited for data-efficient fine-tuning -- otherwise known as \textit{rapid adaptation} -- making the most efficient gradient updates from the few data points available in the target domain.

\section{Fast domain adaptation of a goal-oriented dialogue system}
\label{sec:problem}

\begin{figure*}[ht!]
  \centering
  \includegraphics[width=0.95\linewidth]{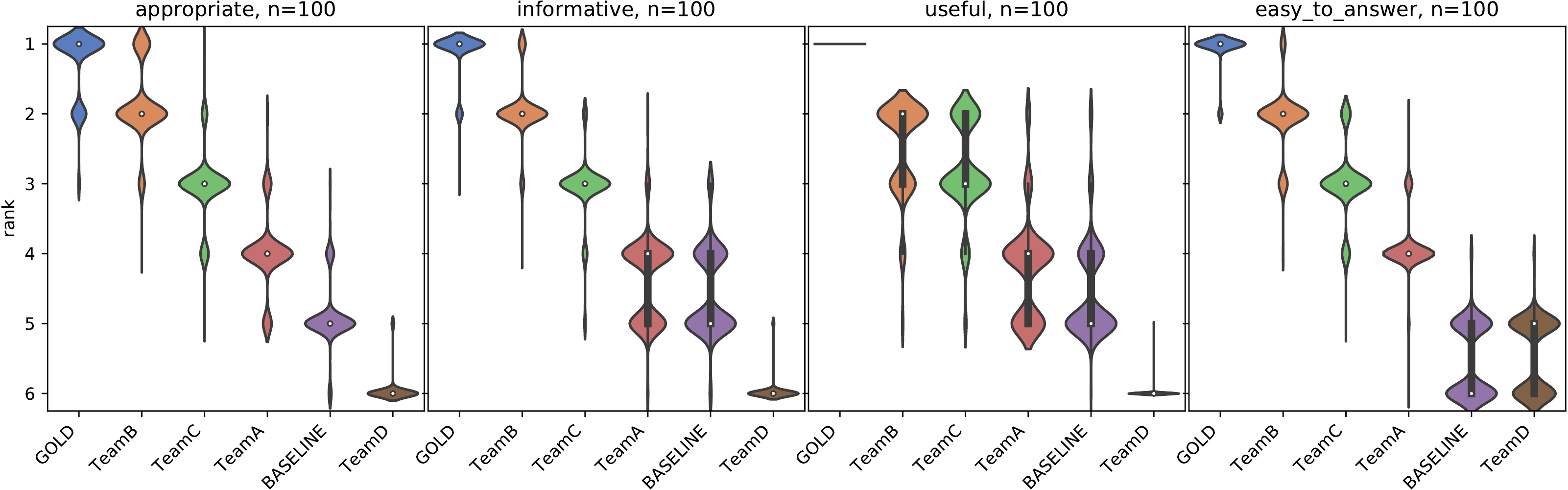}
  \caption{Human evaluation: rank densities by metric with the sample size of 100 dialogues (lower numbers are better). Our submission is denoted as Team B. Densities are determined by drawing 1000 times with replacement from the 100 dialogues and recomputing the rank.}
  \label{fig:human-eval-by-metric}
\end{figure*}

\begin{table}[t!]
  \centering
  \small
    \caption{Ranking from judges' pairwise comparisons}
    \begin{tabularx}{\linewidth}{@{}cXS{}}\toprule
      % & \multicolumn2c{\textbf{Metric}}\\\cmidrule{2-3}
      \textbf{Rank}& \multicolumn1l{\textbf{Submission}}&\multicolumn1c{\textbf{Win rate (\%)}}\\\midrule
      1&Gold response&62.32\\\cmidrule(r{2cm}){1-2}
      \textbf{2}&\textbf{Team B (ours)}&\-\hspace{0pt} \textbf{56.85}\\ % hack with hspace - the cell would get misaligned when bold..
      3&Team C&52.07\\
      4&Team A&47.35\\
      5&Baseline 1&44.18\\
      6&Team D&37.34\\\bottomrule
    \end{tabularx}
    \label{tab:win-rate}
\end{table}

Goal-oriented dialogue systems can be challenging to bootstrap:
For a new domain, little data is available to train e.g. a natural language understanding (\NLU) module or other parts of the pipeline. Often, a Wizard-of-Oz (\WOz, \citealp{kelley84}, \citealp{rieser05}) schema can be used to obtain some initial test data, however, this requires training human agents for the task and setting up a complex pipeline. The value of \WOz data is limited, since ``users'' are mostly hired and might not conform to real users. Additionally, any change in the chatbot interface requires collecting more data.

In the context of the \DSTC[8] domain adaptation challenge, we aim to build a model that predicts user responses for a goal-oriented dialogue system for which only limited in-domain data is available.
Such data could be collected from e.g. customer service transcripts, or written by the developers themselves.
From this in-domain data, the \emph{support set}, we would like to extrapolate responses to novel dialogue contexts (the \emph{target}). However, the support set is typically too small to train a generative dialogue model. Instead, we adapt a generic dialogue model trained on a large corpus of conversations over multiple \emph{source} domains.

Technically, the problem setup is as follows: having trained the base model on the source domains, the model is then fed with one target dialogue and a support set at a time. The model's task is to predict the next user turn of the target dialogue, taking into account the support set before producing a prediction. At prediction time, each target dialogue is processed in isolation from other target dialogues, such that the model cannot use knowledge or state obtained from other target/support data.

\section{Proposed model}
\label{sec:model}

\sisetup{round-mode=places,round-precision=2,per-mode=symbol,detect-all,tight-spacing=true,table-format=2.2}
\newcommand\myhead[1]{\scriptsize\bfseries #1}
\begin{table*}[ht!]
  \small
    \caption{Automatic evaluation results on \metalwoz}
    \begin{tabularx}{\linewidth}{@{}XSSSSSSSSSSSS@{}}
      \toprule
      %\rowfont{\scriptsize}
      &\multicolumn6c{\textbf{\metalwoz pure task (\%)}}&\multicolumn6c{\textbf{\metalwoz cross task (\%)}}\\\cmidrule(r){2-7}\cmidrule(l){8-13}
      &\myhead{BLEU{1}}&\myhead{BLEU{2}}&\myhead{BLEU{3}}&\myhead{CIDEr}&\myhead{METEOR}&\myhead{ROUGE-L}&\myhead{BLEU{1}}&\myhead{\BLEU{2}}&\myhead{BLEU{3}}&\myhead{CIDEr}&\myhead{METEOR}&\myhead{ROUGE-L}\\\midrule
      Retrieval\,\BERT&7.93&4.43&2.87&12.56&7.38&6.91&5.35&2.16&1.05&4.98&4.56&4.52\\
      Retrieval\,\SPFT&9.57&5.37&3.45&14.32&6.98&7.19&5.94&2.25&0.93&4.69&4.29&4.53\\
      \HRED&8.66&3.86&2.11&13.73&6.02&7.75&8.94&3.87&2.02&12.65&6.05&7.55\\\cmidrule(r){1-1}
      \GPT[2] --sup\footnotemark[1]&8.2\phantom0&3.95&2.22&16.41&6.1&8.34&8.37&3.8&2.05&15.6&6.17&8.55\\
      \GPT[2] --ret\footnotemark[2]&11.33&6.45&4.17&23.38&8.23&10.74&10.21&5.26&2.95&18.06&7.06&\bfseries 9.59\\
      \textbf{\GPT[2] hybrid}&\bfseries 12.73&\bfseries7.43&\bfseries4.88&\bfseries28.74&\bfseries9.18&\bfseries11.77&\bfseries10.39&\bfseries5.31&\bfseries2.95&\bfseries18.26&\bfseries7.1&9.27\\\bottomrule
    \end{tabularx}
   \small{\footnotemark[1] does not use support set.  \footnotemark[2] fine-tuned to support set, but does not use retrieval logic} 
    \label{tab:results-metalwoz}
\end{table*}

We use a language model pre-trained on a very large and diverse collection of textual data providing a strong language prior and then adapt the model for our tasks in the form of fine-tuning. Our base model is \GPT[2] \citep{DBLP:journals/corr/abs-1901-08149}, a transformer-based language model.
%Base \GPT[2] model is a large-scale yet general-purpose language model, which we then further train to our dialogue task.
%While base \GPT[2] is trained just from raw text, we have to modify its inputs for dialogue purposes. As such, we use parallel input sequences with (1) the concatenated dialogue context, (2) per-token speaker tags, (3) per-token turn positions in dialogue. All of those are encoded independently, and the resulting representations are then summed up.
In order to adapt \GPT[2] for dialogue generation, we first augment the input embedding for each token in the dialogue with (1) a speaker tag embedding identifying the speaker and (2) a turn embedding, identifying the turn number in the current dialogue. These additional embedding matrices are learned solely using the dialogue data. The input token embeddings are then obtained by summing up these representations. We also add two task-specific output layers (or ``heads'') for our purposes: a language modeling (LM) head and a next-sentence prediction (NSP) classification head, both trained from randomly initialized parameters.

We fine-tune \GPT[2] for response generation by minimizing the negative log-likelihood of response tokens given the concatenation of dialogue context and the previous tokens in the response,
% Language modeling with \GPT[2] architecture works in a Causal Transformer way, i.e. the model is fed with the dialogue context and the generated utterance so far, and it predicts the next token given the concatenated context. It minimizes the Negative Log-Likelihood (NLL) loss:
%
\begin{align}
\label{eq:l_lm}
\begin{split}
\mathcal{L}_{\text{LM}} &= -\log P_{\text{LM}}(X \mid C) \\ &= -\sum_{i=1}^{\left | X \right |}{\log P_{\text{LM}}(x_i \mid x_{i-1}, ..., x_{1}, C)},
\end{split}
\end{align}
where $X$ is the response and $C$ is the dialogue context,~i.e. the concatenation of the tokens in the previous utterances.

To predict the next sentence, we proceed as follows: given a context/response pair $(C, X)$, the classification head is trained to produce a binary label $y$, which is $1$ if $X$ is the correct response given the context $C$, and $0$ if $X$ is a distractor (a random utterance from the corpus). We minimize a binary cross-entropy:
\begin{gather}
\label{eq:l_nsp}
\begin{aligned}
\begin{split}
\mathcal{L}_{\text{NSP}} = & - y \log P_{\text{NSP}}(y \mid X, C) \\
& - (1 - y) \log P_{\text{NSP}}(1 - y \mid X, C),
\end{split}
\end{aligned}\\
\label{eq:nsp}
P_{\text{NSP}}(y \mid X, C) = \text{softmax}(f_{\text{NSP}}(h_{X,C})),
\end{gather}
where $h_{X, C}$ is the last hidden state of the last \GPT[2] layer after having encoded the concatenation of $X$ and $C$ and $f_{\text{NSP}}$ is the next-sentence prediction head (in our case a simple linear transformation). In practice, for each $(C, X)$ pair in the corpus, we sample 1 distractor $\bar X$.

%\begin{equation}
%\label{eq:l_nuc}
%  \begin{split}
%  \mathcal{L}_{\text{NUC}} = - \sum_{i=1}^k{ & y_i \log P(y_i \mid X_i, C_i) \\
%  & + (1 - y_i) \log P(1 - y_i \mid X_i, C_i) },
%  \end{split}
%\end{equation}
%\begin{equation}
%\label{eq:nuc_softmax}
%  P(y \mid X, C) = \text{Softmax}(h_{GPT} * W_{NUC}),
%\end{equation}

%\subsection{Training on source domains}
%\label{sec:training}
We obtain a suitable dialogue prior by fine-tuning the modified \GPT[2] model on the source domains with both the language modeling and next-sentence prediction tasks as described above, therefore minimizing $\mathcal{L} = \mathcal{L}_{\text{NSP}} + \mathcal{L}_{\text{LM}}$.

%The training is organized as follows: we pass into the model dialogue contexts in the form described above together with the `gold' response and a distractor response randomly drawn from the training set. The model's tasks are to generate the gold response with its LM head and tell the gold response and the distractor apart using the NSP head at the same time. MLE losses described above are used as the optimization objectives.

\subsection{Fine-tuning on target domains and prediction}
\label{sec:pred}

As every test dialogue in the target domain/task is accompanied with a small support set of dialogues from the same domain/task, we make use of this data by further fine-tuning the dialogue model on the support dialogues. Crucially, we make sure not to accumulate any information between test dialogues: after each fine-tuning on the support set, we reset the weights of the model to the dialogue prior obtained by the fine-tuning stage described in the previous section.

% The model's state is reverted back to after-training state before each target dialogue to predict, so the model doesn't continuously accumulate fine-tuning progress at this stage.

In order to add diversity to the responses, \GPT[2] uses \textit{nucleus} (top-$p$) sampling \citep{DBLP:journals/corr/abs-1904-09751} during generation, i.e. the model's vocabulary $V$ is pruned into $V^{p}$, the smallest set such that
\begin{align}
  \sum_{x \in V^{p}}{p(x \mid x_{1:i-1}, C)} \ge p,
\end{align}
and the final distribution from which the words are sampled is rescaled as follows:
\begin{align}
  P'(x \mid x_{1:i-1}) = 
  \begin{cases}
      \frac{P(x \mid x_{1:i-1}, C)}{\sum_{x \in V^{p}}{P(x \mid x_{1:i-1}, C)}}& \text{if } x \in V^{(p)}\\
      0,  & \text{otherwise.}
  \end{cases}
\end{align}

\subsection{Hybrid generative-retrieval prediction}
\label{sec:hybrid}

In our experiments, we found that retrieval baselines are quite effective in the automatic metrics considered. Combining retrieval techniques with our generative model in a hybrid approach produced a stronger model.

The retrieval component is set up as follows: when predicting the $t$-th turn of the test dialogue, the model embeds its context of length $t-1$ as well as all the support dialogue contexts of the same length $t-1$ using the fine-tuned dialogue encoder. The encoding for the dialogue context is the hidden state of the last layer of the Transformer model at the position corresponding to the last token in the context. Then, it selects the nearest support context to the target context and picks its $t$-th turn as the retrieved candidate response.

Finally, the model's own generated response and the best retrieved candidate response are ranked using the NSP classification head,~i.e.\ both responses are concatenated with the ground-truth context and the one with the higher $P_{\text{NSP}}$ (Eq.~\ref{eq:nsp}) is selected. The model is visualized in Figure~\ref{fig:model}.

\vspace{5mm plus 5mm minus 5mm}
\section{Baselines}
\label{sec:baselines}

\begin{table}[t!]
  \centering
  \small
    \caption{Automatic evaluation results on \multiwoz pure task dataset}
    \begin{tabularx}{\columnwidth}{@{}XSS{}}\toprule
      % & \multicolumn2c{\textbf{Metric}}\\\cmidrule{2-3}
      & \multicolumn1c{\myhead{Intent F1 (\%)}} & \multicolumn1c{\myhead{Intent+Slots F1 (\%)}}\\\midrule
      Retrieval \BERT&48.00&21.95\\
      Retrieval \SPFT&51.53&26.58\\
      \HRED&44.61&35.57\\\cmidrule(r{12mm}){1-1}
      \GPT[2] --sup\footnotemark[1]&58.54&43.07\\
      \GPT[2] --ret\footnotemark[2]&48.00&37.24\\\cmidrule(r{12mm}){1-1}
      Team D&    54.98&42.34\\
      Team C&    61.40&41.87   \\
      \GPT[2] hybrid (Team B)&64.50&48.33\\
      \bfseries Team A&\bfseries 78.70  &   \bfseries 60.00\\\bottomrule
    \end{tabularx}
    \label{tab:results-multiwoz}
   \vspace*{15mm plus 5mm minus 5mm}
\end{table}

% this table is not very informative, should we just include all rankings? copy from our track overview paper?
%\begin{table*}[ht!]
%    \centering
%    \caption{Human evaluation results on \metalwoz held-out dataset}
%    \begin{tabularx}{\linewidth}{@{}Xc@{\;\;}c@{\;\;}c@{\;\;}c@{\;\;}c@{\;\;}c@{\;\;}c@{}}
%    \toprule
%        & \multirow{2}{*}{\textbf{Overall}}  &  \multicolumn2c{\textbf{Testset}} & \multicolumn4c{\textbf{Metric}} %\\\cmidrule(r){3-4}\cmidrule(r){5-8}
%        && \multicolumn1c{Pure}  & \multicolumn1c{Cross}  & \multicolumn1c{Appropriate}  & \multicolumn1c{Easy to Answer}  & \multicolumn1c{Informative}  & \multicolumn1c{Useful} \\\midrule
%        Retrieval SP+FT & 4 & 4 & 4 & 4 & 5 & 4 & 4 \\
%        \textbf{\GPT[2] hybrid} & 1 & 1 & 1 & 1 & 1 & 1 & 1 \\
%    \bottomrule
%    \end{tabularx}
%\end{table*}

We compare our hybrid model to the retrieval baselines provided by the \DSTC[8] organizers. The baselines ignore the training data and rely solely on the support sets: they embed each support dialogue's context and find the one nearest to the target context using cosine distance as the metric. They then return the turn following the identified context as the predicted response. There are two baselines, which differ in their encoder: (1) \BERT \citep{DBLP:conf/naacl/DevlinCLT19}-based, taken off-the-shelf, and (2) SentencePiece/FastText-based, modeled after \citet{DBLP:conf/emnlp/GuWCLC18} with embeddings pre-trained on the Reddit Conversations corpus.

We also compare our model to a bidirectional \LSTM-based \HRED~\citep{DBLP:conf/aaai/SerbanSBCP16} trained on \metalwoz. Given the time constraints, we could only evaluate a base model without fine-tuning to support sets.

\section{Datasets}
\label{sec:data}

\begin{table*}[ht!]
\small
\center
\caption{GPT-2 Hybrid example responses}
\begin{tabular}{@{}rll@{}}\toprule

\textbf{Context}&\texttt{Wiz}&Hello how may I help you?\\
&\texttt{Usr}&I need some advice about Rome.\\
&\texttt{Wiz}&What do you want to know?\\
\textbf{Gold response}&\texttt{Usr}&I like history, where should I go?\\
\textbf{Predicted response}&\texttt{Usr}&Who was the founder of Rome?\\
\hline

\textbf{Context}&\texttt{Wiz}&Hello, how may I help you?\\
&\texttt{Usr}&I'm looking for a train to broxbourne.\\
&\texttt{Wiz}&Where would you like to leave from, and what day and time would you like to travel?\\
&\texttt{Usr}&I'm going to be leaving from Cambridge. I need to leave on Tuesday by 13:00.\\
&\texttt{Wiz}&Is there a time you would like to get there by?\\
\textbf{Gold response}&\texttt{Usr}&I don't have an arrival preference.\\
\textbf{Predicted response}&\texttt{Usr}&No I don't have a leave time preference. Just not too early if I don't have too.\\
\hline

\textbf{Context}&\texttt{Wiz}&Hello how may I help you?\\
&\texttt{Usr}&I'd like to book a flight.\\
&\texttt{Wiz}&Okay, any particular place in mind?\\
&\texttt{Usr}&How am I supposed to book a flight to Greece? Me and my kawaii girlfriend were wondering.\\
&&She says ``Ooooo, Greece. :3''\\
&\texttt{Wiz}&I can book a flight for two if you want, There will be an evening flight\\
&\texttt{Usr}&Hmm, wait, I don't really wanna book a flight. We were just curious! She says ``Hey! No bookies! :(''\\
&\texttt{Wiz}&Oh, I was confused by that. Can I have your email address so I could send several flight options for you\\
\textbf{Gold response}&\texttt{Usr}&Sure thing! My email address is ``weeabooking@otaku.corn''. She says ``I wanna watch my anime now! ;\_;''\\
\textbf{Predicted response}&\texttt{Usr}&Well, I guess I'll just get back to you. Thanks!\\
\bottomrule

\end{tabular}
\label{tab:examples}
\end{table*}

We use \metalwoz, the dataset for \DSTC[8] Track~2 ``Fast Domain Adaptation'' \citep{lee2019multi-domain}. It contains more than 37,000 human-human dialogues spanning the total of 227 tasks in 47 domains. The dialogues are collected in a \textit{Wizard-of-Oz} style: human participants were assigned the role of bot or user, then given a problem domain and related specific task, and instructed to reach the user's goal over at least 10 dialogue turns.

For evaluation purposes, we additionally use \multiwoz~\cite{DBLP:conf/emnlp/BudzianowskiWTC18}, another multi-domain, multi-task dialogue dataset. Dialogues in \multiwoz contain \NLU annotations, particularly for intent and slots, which we use in order to to evaluate the systems' goal-oriented performance. A subset of \multiwoz{} (\multiwoz{} pure task), where dialogues only pertain to a single domain, was used for evaluation.

\section{Experimental setup and evaluation}
\label{sec:setup}

\begin{table}[ht!]
  \centering
  \small
    \caption{\GPT[2]-hybrid generate/retrieve response ratio}
    \begin{tabularx}{\linewidth}{@{}XSS{}}\toprule
      % & \multicolumn2c{\textbf{Metric}}\\\cmidrule{2-3}
      \textbf{Dataset / domain}& \multicolumn1c{\textbf{Generated (\%)}}&\multicolumn1c{\textbf{Retrieved (\%)}}\\\midrule
      \textbf{\metalwoz pure task}&\\
      \-\hspace{5pt} booking flight&64.1&35.9\\
      \-\hspace{5pt} hotel reserve&63.8&36.2\\
      \-\hspace{5pt} tourism&57.4&42.6\\
      \-\hspace{5pt} vacation ideas&61.7&38.3\\%\midrule
      \textbf{\metalwoz cross task}&\\
      \-\hspace{5pt} booking flight&68.2&31.8\\
      \-\hspace{5pt} hotel reserve&74.8&25.2\\
      \-\hspace{5pt} tourism&73.9&26.1\\
      \-\hspace{5pt} vacation ideas&74.7&25.3\\%\midrule
      \textbf{\multiwoz}&\\
      \-\hspace{5pt} attraction&55.6&44.4\\
      \-\hspace{5pt} hospital&60.0&40.0\\
      \-\hspace{5pt} hotel&63.0&37.0\\
      \-\hspace{5pt} police&52.1&47.9\\
      \-\hspace{5pt} restaurant&61.3&38.7\\
      \-\hspace{5pt} taxi&64.3&35.7\\
      \-\hspace{5pt} train&61.0&39.0\\\bottomrule
    \end{tabularx}
    \label{tab:gen-ret-ratios}
\end{table}

We perform training in two stages: training of the base model and fine-tuning it to the target dialogue's support set.
At the first stage, we train the model for the maximum of 5 epochs with early stopping. The second stage goes on for 1 epoch in the interest of time. \GPT[2] models use the context of 3 exchanges, or 5 turns: bot-user-bot-user-bot, predicting the next user's utterance. We mainly used the `small' \GPT[2] checkpoint by HuggingFace~---we also tried the `medium' one, but found no improvement with it in our task. % We train our models on 4 NVidia V100 GPUs, with prediction+fine-tuning stage distributed per-dataset per-domain, consuming up to 60 GPUs in total.

\subsection{Human evaluation}

The main systems' goal is to generate appropriate responses towards maintaining a natural cooperative dialogue on the user's side, so the main evaluation is performed involving human judges. Specifically, Amazon Mechanical Turk workers were tasked to compare the candidate responses given the dialogue context. Each comparison was pairwise between the results of two systems presented in random order. Judges ranked the responses against the following criteria:
\begin{itemize}
  \item \textit{Usefulness}~--- whether the response is useful given the dialogue context and the user's overall final goal,
  \item \textit{Informativeness}~--- whether the response specifically contains information relevant to the conversation,
  \item \textit{Appropriateness}~--- whether the response is appropriate (on-topic, of a reasonable length, not repetitive) to the conversation,
  \item \textit{Easiness to answer}~--- given a hypothetical conversational bot on the system side, whether the response will be a valid input for it and presumably straightforward to process.
\end{itemize}

For each pairing, 3 independent comparisons were performed against each metric. 
The number of comparisons required was reduced by letting the \emph{Multisort} algorithm \citep{maystre2017just} determine which responses to compare, causing more similar systems with similar performance to be compared more often with each other. Bootstrapping over the 100 randomly chosen dialogue contexts was used to determine average ranks and assess the ranking robustness \citep{hall2009using}.

% \vspace*{5mm plus 5mm}
\subsection{Automatic evaluation}

In addition to human evaluation, we also assess  model performance using automatic metrics. The models were evaluated on \metalwoz against word-overlap metrics such as \BLEU-1--3, \CIDEr, \METEOR, \ROUGEL using the \NLGEval package \citep{sharma2017nlgeval}. Although not ideal for the specifics of dialogue and spoken language in general \citep{adem17,dziri19-evaluating}, such metrics approximate the overall quality of a generative model and are especially useful for intermediate evaluation. We evaluate models in two modes on \metalwoz: in \textit{pure task}, support dialogues are drawn from the same domain and task as target dialogue; in \textit{cross-task}, support and target dialogues are from the same domain, but different tasks.

We also perform additional evaluation of Entity/Intent F1 of the \multiwoz dataset in pure task mode with pre-trained \NLU taggers from the ConvLab package \citep{lee2019convlab}. There is no \multiwoz data available at the first stage (base model training), so all the exposure our model has to this dataset is via support dialogues. Complementary to \metalwoz evaluation, this stage is designed for assessing the models' goal-oriented performance.

\section{Results and discussion}
\label{sec:results}

\subsection{Human evaluation}
Results of pairwise comparisons are shown in Table~\ref{tab:win-rate}. Our \GPT[2] hybrid system's responses (Team~B) were preferred by the judges in \textbf{56\%} of direct comparisons. This surpasses the next best system (Team C) performance by more than \textbf{4\%}, with only the gold human responses being chosen more frequently.

Furthermore, from the bootstrap ranking distribution (Figure~\ref{fig:human-eval-by-metric}), we see that, apart from the gold human responses, our model's outputs are consistently preferred over other submissions by the judges. Of all metrics used, the most notable are `appropriateness' and `usefulness'. On the former, \GPT[2] hybrid's responses have the second visible peak at rank 1 competing with gold responses. On usefulness however, rank~1 is held by the gold responses with no variation, and our model has the second visible peak at rank 3, thus almost tying with Team C.

\subsection{Automatic evaluation}

Results on \metalwoz and \multiwoz against automatic evaluation metrics are shown in Tables \ref{tab:results-metalwoz} and \ref{tab:results-multiwoz}, respectively. We observe that retrieval baselines attain very competitive performance on both datasets, with FastText embeddings from Reddit leading to overall better results than off-the-shelf \BERT, especially in the \textit{pure task} setting.

With \GPT[2], we performed an ablation study to have a closer look into its performance. We evaluated three versions: `hybrid' which we presented in this paper, `--ret' with retrieval logic turned off, and `--sup' with no retrieval logic and no fine-tuning to the support set. As seen in the Table \ref{tab:results-metalwoz}, there is strong dependence on support dialogues (`--sup' vs\@. `--ret') as the base model mostly struggles to compete with the baselines. Adding retrieval logic (`hybrid' vs\@.\ `--ret') results in further performance gains. \HRED and \GPT[2]--sup, the two models that did not use support dialogues, had comparable performance on \metalwoz.

In goal-oriented metrics on \multiwoz (see Table \ref{tab:results-multiwoz}), the same performance pattern is observed with retrieval models, but \GPT[2] in the generative-only version performs surprisingly better when not fine-tuned to support set (`--sup'). On the other hand, the hybrid model experiences even more performance gain than on \metalwoz. Presumably, generating responses for this dataset is harder due to the fact that it is not represented at the main training stage, and there is not much utterance overlap with \metalwoz, so little knowledge transfer takes place in this experiment. Compared to other submissions, we observe that \GPT[2] hybrid still outperforms most of the competitors and only gives way to Team A's system. We hypothesize here the best \multiwoz{} model (Team A) was fitted to the automatic evaluation metrics too tightly, with the negative side effect observable in human evaluation results of Table \ref{tab:win-rate} and Figure \ref{fig:human-eval-by-metric}, where this system was prevalently ranked 4${}^{\text{th}}$ and 5${}^{\text{th}}$.

\paragraph{Retrieval and Generation Frequency}
In Table~\ref{tab:gen-ret-ratios}, we show per-domain ratios of re\-trie\-ved/ge\-n\-e\-ra\-ted responses from the hybrid model. We find that the majority of the responses are generated, and the retrieval logic works as the fallback option. On \metalwoz, which the model had more exposure to during the training, generated responses ratio is generally slightly higher than that on \multiwoz which was only seen by the model via support dialogues. Consequently, the model's overall confidence on this dataset is lower, which results in more frequent fallbacks.

Overall, we observe in Table~\ref{tab:examples} that there are many cases in the data where the gold response cannot possibly be inferred from the dialogue context. Specifically, the task was posed in the way that no extra data, such as a  knowledge base or task description, was provided to the system~--- therefore, the main goal intended for the hypothetical ideal system is to naturally model human responses in a co-operative goal-oriented dialogue, and to do that in a data-efficient way. This is reflected in the way human judges are asked about response quality.

\section{Conclusion and future work}
\label{sec:future}

We presented a hybrid generative/retrieval approach to goal-oriented dialogue with fast domain adaptation via transfer learning. It attains robust and diverse language generation performance across domains, and uses retrieval logic as a fallback mechanism in cases of low confidence. Our method is the winning entry at the \DSTC[8] Fast Domain Adaptation task achieving state-of-the-art performance as evaluated with human judges. In additional automatic evaluation, it attains competitive generalization performance in adaptation to the goal-oriented \multiwoz dataset without any exposure to that data during the main training stage.

Overall, we observe that transfer learning, while being in the core of state-of-the-art methods for dialogue domain adaptation and few-shot learning \cite{DBLP:journals/corr/abs-1910-01302,DBLP:journals/corr/abs-1908-05854}, still does not attain the performance level sufficient for direct adoption in industry. It's evident that the problem of data-efficient dialogue response generation needs further research, and one promising direction that we are going to explore in our own future work is the meta-learning framework \cite{DBLP:conf/acl/QianY19}, or `learning to fine-tune'. Based on splitting the task into multiple subtasks and solving them with separate versions of the model with further merging of each individual learner's progress, meta-learning approach will naturally fit our multi-domain setup as well as lead to potentially better fine-tuning performance.

\bibliographystyle{aaai}
\bibliography{refs.bib}

\end{document}